\documentclass[runningheads]{llncs}
\usepackage{graphicx}
\usepackage{float}

\begin{document}

\title{Amortised Design Optimization for Item Response Theory}

\titlerunning{Amortised Design Optimization for IRT}

\author{Antti Keurulainen\textsuperscript{1,2}, Isak Westerlund\textsuperscript{2}, Oskar Keurulainen\textsuperscript{2}, Andrew Howes\textsuperscript{1,3}}

\authorrunning{Keurulainen et al.}

\institute{Aalto University, Espoo Finland \and
Bitville Oy, Espoo Finland \and
University of Birmingham, United Kingdom}

\maketitle

\begin{abstract}
Item Response Theory (IRT) is a well known method for assessing responses from humans in education and psychology. In education, IRT is used to infer student abilities and characteristics of test items from student responses. Interactions with students are expensive, calling for methods that efficiently gather information for inferring  student abilities. Methods based on Optimal Experimental Design (OED) are computationally costly, making them inapplicable for interactive applications. In response, we propose incorporating amortised experimental design into IRT. Here, the computational cost is shifted to a precomputing phase by training a Deep Reinforcement Learning (DRL) agent with synthetic data. The agent is trained to select optimally informative test items for the distribution of students, and to conduct amortised inference conditioned on the experiment outcomes. During  deployment the agent estimates parameters from data, and suggests the next test item for the student, in close to real-time, by taking into account the history of experiments and outcomes.

\keywords{Item Response Theory (IRT)  \and Experimental Design \and Deep Reinforcement Learning (DRL).}
\end{abstract}

\section{Introduction}

Item Response Theory (IRT) is a method for inferring student abilities and test item characteristics by observing test outcomes conducted with students \cite{hambleton2013item}. In IRT, the relationship between an individual's ability and their response to a test item is typically modeled with logistic regression, describing the probability that an individual with a certain ability will give a correct response to a test item with a given difficulty. Inferring student abilities and test item characteristics offers benefits such as user skill assessment, test item calibration, and optimization of learning experiences for students by Intelligent Tutoring Systems (ITS) or human tutors.

To fully exploit the benefits of IRT in real-time interactions with students, each interaction should provide as much information as possible. Optimal Experimental Design (OED) aims to select an experiment that maximizes an information criterion \cite{ryan2016review,smucker2018optimal}. Combining OED with IRT enables designing experiments that are maximally informative about student abilities.

We incorporate amortised experimental design into IRT and construct a system in which the next test for the student, conditioned on the previous test outcomes, can be given in near-real time. Our approach involves training a Deep Reinforcement Learning (DRL) agent, which allows for amortising both the OED and parameter estimation tasks simultaneously.

\section{Background and related work}

The most simple IRT variant is the 1PL model, also known as the Rasch model \cite{rasch1993probabilistic}. The probability that a student $i$ gives a correct answer for the item $j$ is defined as:
\begin{equation}
P(y_{i,j} = 1|\theta_i, b_j) = \frac{1}{1+e^{-{(\theta_i - b_j)}}}
\label{eq1}
\end{equation}
where $y_{i,j}$ is the Bernoulli-distributed outcome, $\theta_i$ is the ability of student $i$, and $b_j$ is the difficulty of the item $j$. The task is to infer the parameters $\theta$ and $b$ from the observations, i.e., from the test outcomes. We consider a setting where only the student ability $\theta$ is estimated, while the item parameter $b$ is treated as a design parameter. This formulation models an experimental design setting where items can be selected for students to obtain information about their latent abilities.

Recent work has focused on inferring IRT parameters using deep neural network variants. Wu et. al. \cite{wu2020variational} suggest IRT inference using Variational Inference (VI). Paassen et. al \cite{paassen2022sparse} combine sparse factor analysis with Variational Autoencoders (VA). Deep-IRT \cite{yeung2019deep} combines key-value networks with IRT to improve the explainability of deep learning-based knowledge tracing. Closely related to IRT is Knowledge Tracing (KT), which tracks and models students' knowledge over time based on their responses during a set of interactions \cite{corbett1994knowledge}. Li et al. \cite{li2020deep} suggest a value-based RL method for an adaptive learning system for optimal teaching. Ghosh et al. \cite{ghosh2021bobcat} suggest BOBCAT, casting  experimental design for IRT as a metalearning problem. However, their question selection algorithm assumes questions to be discrete variables. 

Advances in Deep Learning have opened up new possibilities for OED. It allows for amortising the OED process by pretraining a deep neural network before deployment using existing or synthetic data. During execution, a pretrained neural network can potentially generalize to unseen models within the prior distribution used in the amortisation stage. A deep neural network can also be conditioned on the outcomes of previous interactions for learning an adaptive policy where all previous interactions affect the selection of the next design. However, optimizing the exact mutual information directly might lead to an intractable solution because of the need to calculate nested expectations \cite{rainforth2018nesting}. Recently, various suggestions have been made to solve these challenges. Foster et. al. suggest Deep Adaptive Design (DAD) \cite{foster2021deep}, which provides non-myopic, amortised experimental design by maximising an approximate lower bound of the mutual information instead of an exact mutual information objective. Blau et. al. \cite{blau2022optimizing} introduce an RL-based method, highlighting the strong exploration and exploitation tradeoff capability.

\section{Amortised Design Optimisation for IRT (ADOIRT)}

In this section, we propose an amortised design optimisation approach to IRT, which we name Amortised Design Optimisation for IRT (ADOIRT).

\subsection{POMDP formulation for ADOIRT}

We formulate design selection and parameter estimation as a Partially Observable Markov Decision Process (POMDP) given by a tuple $\langle S, A, T, R, O, f, \rho_0, \gamma \rangle$, and find an approximately optimal policy using Reinforcement Learning. The agent selects actions $a_t = (d_t, \hat{\theta_t}) \in A$, where $A$ is the action space consisting of all possible actions, including design requests and estimated student abilities.

At each time step $t$, the environment is in state $s_t \in S$, where $S$ is the state space consisting of all possible states. The state $s_t = \theta_t$ is the true student ability. The transition function $T: S \times A \times S \rightarrow [0,1]$ specifies deterministic dynamics: $p(\theta_{t+1} | a_t, \theta_t) = 1$ if $\theta_{t+1} = \theta_t$ and $0$ otherwise. The initial state $s_0$ is sampled from the prior distribution $\rho_0 = p(\theta)$, and $\gamma$ is the discount factor.

Observations are sampled from the observation function $f: S \times A \rightarrow O$ such that $o_t = (\hat{d_t}, y_t) \sim f(s_t,a_t)$. The observation function consists of two parts: mapping the requested design $d_t$ to a corrupted item $\hat{d_t}$, and obtaining the experiment outcome $y_t$. The corrupted item results from selecting the best item during deployment or adding Gaussian noise during training. The outcome is given by $y_t \sim p(y_t|\theta_t, \hat{d_t})$.

The reward function $R: S \times A \rightarrow \mathcal{R}$ specifies the reward received by the agent for taking a particular action in a particular state. The reward is the squared error between the true and estimated student ability: $R(s_t, a_t) = (\theta_t - \hat{\theta_t})^2$. Consequently, the agent is rewarded for selecting experiments that lead to accurate parameter estimations.

\subsection{ADOIRT Architecture and training}

\begin{figure}
\includegraphics[width=1.0\textwidth]{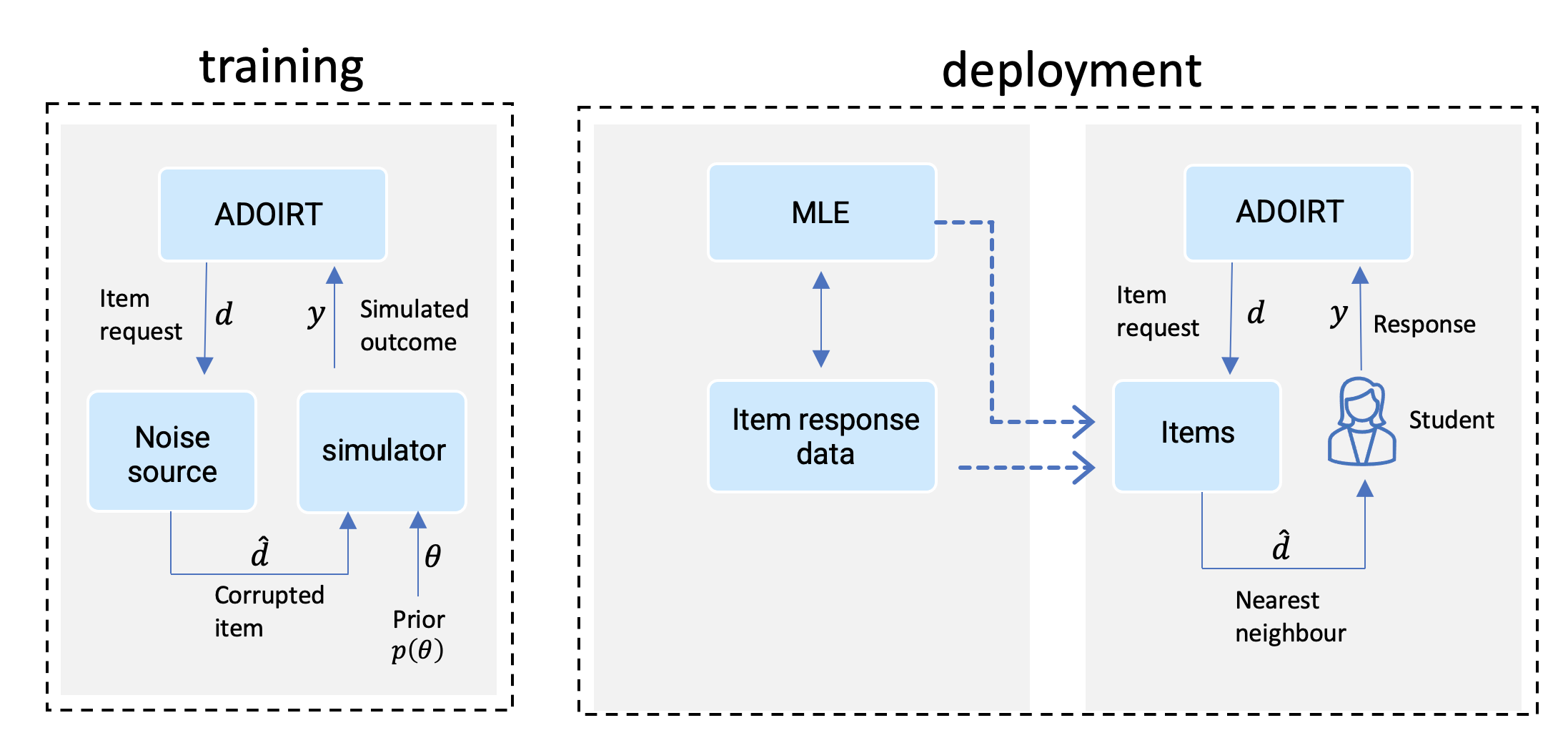}
  \caption{ADOIRT architecture. 
Left panel (training): 
During training, synthetic data is generated by sampling the student abilities and item difficulties from a prior distribution. The simulator uses the obtained item parameters and the IRT model to produce outcomes. Right panel (deployment): In the deployment phase, the item parameters can be estimated beforehand with a standard method, such as MLE, using any existing data. The item requests by ADOIRT are mapped to the closest estimated item difficulty in the collection of available items.}
  \label{fig:ADOIRT}
\end{figure}

The training architecture is illustrated in the left panel of Figure \ref{fig:ADOIRT}. At the start of each episode, the simulator is initialised by sampling new model parameters from the prior. ADOIRT makes requests for the item difficulties, and the simulator produces an outcome using the obtained design and sampled student ability. The outcome and obtained design are concatenated to the history of observations.

In the deployment phase, the trained ADOIRT can be used with existing item response data. First, a MLE estimate for the item difficulties is produced using stochastic gradient descent (SGD). Once the estimates of the item difficulties are available, the item requests by the ADOIRT are mapped to the closest items from the existing dataset. Based on the outcomes of the student, the goal of the agent is to select informative items from the existing collection of items.

The policy network implementation is a multilayer perceptron (MLP) network with four layers followed by mean pooling across experiment trials, an output layer, and separate heads for action distribution and value estimation, which are two-layer MLP networks. ReLU activation functions are used after each MLP layer. We train the policy function with the Proximal Policy Optimisation (PPO) algorithm \cite{schulman2017proximal} using the Stable Baselines 3 (SB3) library \cite{stable-baselines3}.

\section{Results}

We assess the performance of ADOIRT in estimating student abilities in a scenario where the unknown item difficulties are inferred using maximum likelihood estimation. We simulate synthetic datasets with 200 students and 50 items for training and evaluation. The item request by the agent is then mapped to the closest estimated difficulty in this collection.

We compare the performance with a situation where design values are randomly chosen. To gain further insight into the performance, we also trained the agent by concealing experiment outcomes from the observation until the final time step of the episode. In this case, the agent learned a well performing, non-adaptive design strategy.

\begin{figure}
\includegraphics[width=1.0\textwidth]{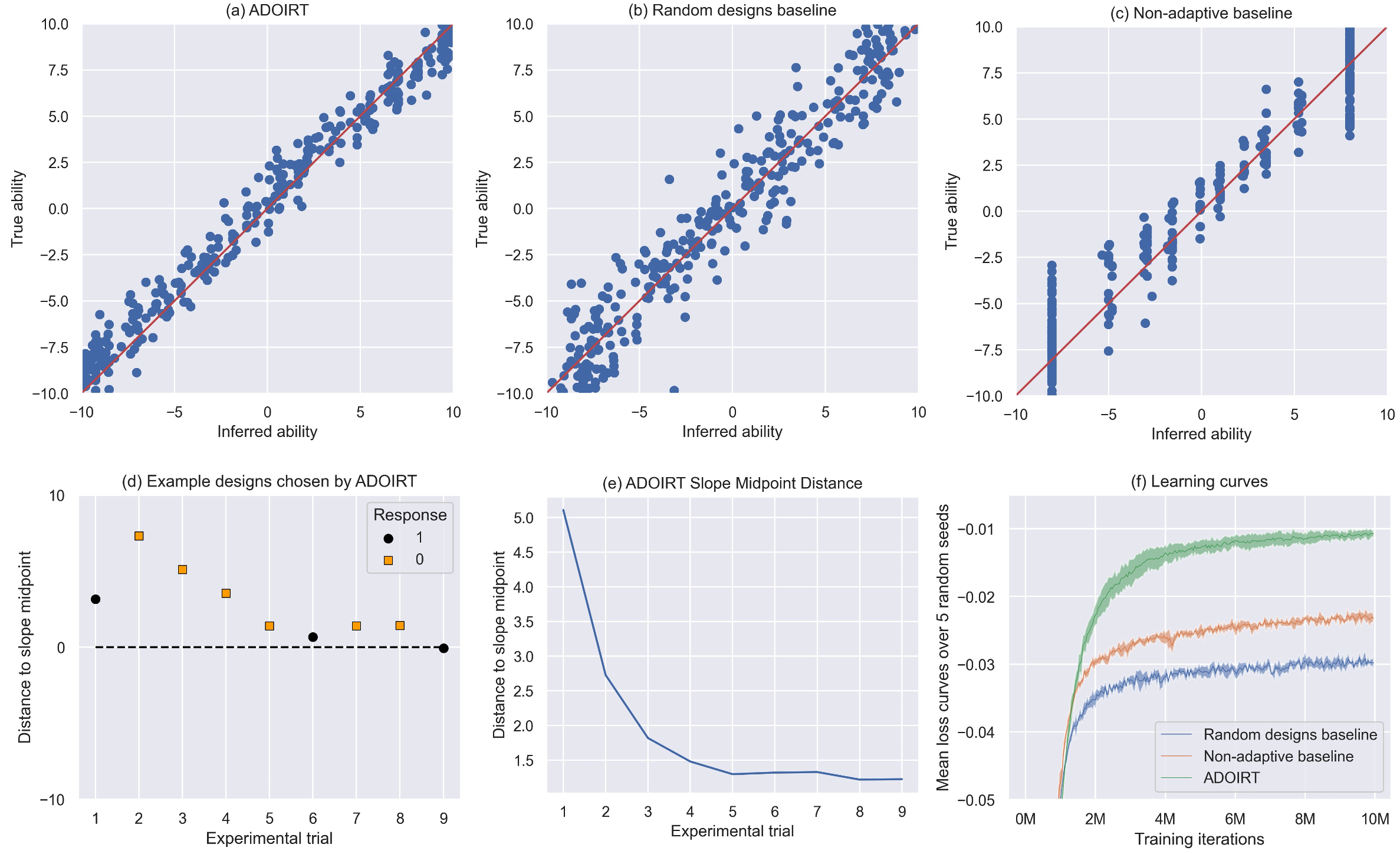}
  \caption{Panel (a): True vs inferred ability for ADOIRT. Panel (b): True vs inferred ability for random designs. Panel (c): True vs inferred ability for non-adaptive optimized designs. Panel (d): An example case where 10 items are presented to the student and the agent successfully converges on designs close to the midpoint of the sigmoid function.  Panel (e): ADOIRT's design values converging to sigmoid midpoint (1000 models). Panel (f): Agent training learning curves with 1 std shaded area. The shaded area represents 1 std over 5 training runs. Panels (a)-(e) use the lowest training error seed.}
   
  \label{fig:study2}
\end{figure}

Figure \ref{fig:study2} illustrates the key results. It shows that experiments chosen by ADOIRT result in lower error in student ability estimation compared to the baselines (panels a-c). In the case of non-adaptive designs (panel c), the inferred student abilities are clustered in ten clusters. This is a reasonable non-adaptive design strategy, as it effectively covers the design space with a limited number of points. Furthermore, the clusters are denser closer to zero, as incorrect predictions for student abilities far from the mean lead to higher penalties. The quantitative results are presented in Table \ref{table2}.

\begin{table}[H]
\caption{ADOIRT-based inference surpasses non-adaptive and random designs in inferring student abilities. This is evaluated using 50 datasets, MLE estimates, and averaging performance over 1000 episodes per dataset. Mean and standard error are computed by repeating the training with five random seeds.}
\begin{tabular}{l|c|c}
\hline
Method & MSE mean & MSE standard error \\ 
\hline
\textbf{Inference with ADOIRT} & \textbf{1.91} & \textbf{0.03} \\
Inference with non-adaptive designs & 3.05 & 0.07 \\
Inference with random designs & 3.39 & 0.06 \\
\hline
\end{tabular}
\label{table2}
\end{table}

\section{Discussion and further work}

In this article, we introduced a novel method for amortised experimental design and parameter estimation for the IRT setting, named ADOIRT. Our studies showed that ADOIRT outperforms non-adaptive and random design strategies, thus being capable of inferring student abilities from a small number of interactions in near real-time. Interesting future work would involve testing the system in a real-world setting with human participants.

\bibliographystyle{splncs04}
\bibliography{aied23}

\begin{thebibliography}{10}
\providecommand{\url}[1]{\texttt{#1}}
\providecommand{\urlprefix}{URL }
\providecommand{\doi}[1]{https://doi.org/#1}

\bibitem{blau2022optimizing}
Blau, T., Bonilla, E.V., Chades, I., Dezfouli, A.: Optimizing sequential
  experimental design with deep reinforcement learning. In: International
  Conference on Machine Learning. pp. 2107--2128. PMLR (2022)

\bibitem{corbett1994knowledge}
Corbett, A.T., Anderson, J.R.: Knowledge tracing: Modeling the acquisition of
  procedural knowledge. User modeling and user-adapted interaction
  \textbf{4}(4),  253--278 (1994)

\bibitem{foster2021deep}
Foster, A., Ivanova, D.R., Malik, I., Rainforth, T.: Deep adaptive design:
  Amortizing sequential bayesian experimental design. In: International
  Conference on Machine Learning. pp. 3384--3395. PMLR (2021)

\bibitem{ghosh2021bobcat}
Ghosh, A., Lan, A.: Bobcat: Bilevel optimization-based computerized adaptive
  testing. arXiv preprint arXiv:2108.07386  (2021)

\bibitem{hambleton2013item}
Hambleton, R.K., Swaminathan, H.: Item response theory: Principles and
  applications. Springer Science \& Business Media (2013)

\bibitem{li2020deep}
Li, X., Xu, H., Zhang, J., Chang, H.h.: Deep reinforcement learning for
  adaptive learning systems. Journal of Educational and Behavioral Statistics
  p. 10769986221129847 (2020)

\bibitem{paassen2022sparse}
Paa{\ss}en, B., Dywel, M., Fleckenstein, M., Pinkwart, N.: Sparse factor
  autoencoders for item response theory. In: Proceedings of the 15th
  International Conference on Educational Data Mining. p.~17 (2022)

\bibitem{stable-baselines3}
Raffin, A., Hill, A., Gleave, A., Kanervisto, A., Ernestus, M., Dormann, N.:
  Stable-baselines3: Reliable reinforcement learning implementations. Journal
  of Machine Learning Research  \textbf{22}(268), ~1--8 (2021),
  \url{http://jmlr.org/papers/v22/20-1364.html}

\bibitem{rainforth2018nesting}
Rainforth, T., Cornish, R., Yang, H., Warrington, A., Wood, F.: On nesting
  monte carlo estimators. In: International Conference on Machine Learning. pp.
  4267--4276. PMLR (2018)

\bibitem{rasch1993probabilistic}
Rasch, G.: Probabilistic models for some intelligence and attainment tests.
  ERIC (1993)

\bibitem{ryan2016review}
Ryan, E.G., Drovandi, C.C., McGree, J.M., Pettitt, A.N.: A review of modern
  computational algorithms for bayesian optimal design. International
  Statistical Review  \textbf{84}(1),  128--154 (2016)

\bibitem{schulman2017proximal}
Schulman, J., Wolski, F., Dhariwal, P., Radford, A., Klimov, O.: Proximal
  policy optimization algorithms. arXiv preprint arXiv:1707.06347  (2017)

\bibitem{smucker2018optimal}
Smucker, B., Krzywinski, M., Altman, N.: Optimal experimental design. Nat.
  Methods  \textbf{15}(8),  559--560 (2018)

\bibitem{wu2020variational}
Wu, M., Davis, R.L., Domingue, B.W., Piech, C., Goodman, N.: Variational item
  response theory: Fast, accurate, and expressive. arXiv preprint
  arXiv:2002.00276  (2020)

\bibitem{yeung2019deep}
Yeung, C.K.: Deep-irt: Make deep learning based knowledge tracing explainable
  using item response theory. arXiv preprint arXiv:1904.11738  (2019)

\end{thebibliography}

\end{document}